\documentclass[fleqn]{article}
\usepackage{graphicx}
\usepackage{amsmath}
\usepackage{amssymb}
\usepackage{amsfonts}
\usepackage{color}

\newcommand{\comment}[1]{{}}

\begin{document}
\bibliographystyle{unsrt}

\title{Anticipating epileptic seizures through the analysis of EEG synchronization as a data classification problem}

\author{
Paolo Detti \thanks{Dipartimento di
Ingegneria dell'Informazione e Scienze Matematiche, Universit\`a di Siena, Via Roma 56, 53100 Italy, e-mail detti@dii.unisi.it, tel.: +39 0577-234850 (1022), fax: +39 0577-233602} \and Garazi Zabalo Manrique de Lara \thanks{Dipartimento di
Ingegneria dell'Informazione e Scienze Matematiche, Universit\`a di Siena, Via Roma 56, 53100 Italy, e-mail garazizml@gmail.com} 
 \and Renato Bruni \thanks{Dip. di
Ingegneria Informatica, Automatica e Gestionale, Universit\`a Sapienza di Roma, Via Ariosto 25, 00185 Italy, e-mail bruni@dis.uniroma1.it}
 \and Marco Pranzo \thanks{Dipartimento di
Ingegneria dell'Informazione e Scienze Matematiche, Universit\`a di Siena, Via Roma 56, 53100 Italy, e-mail pranzo@dii.unisi.it}
 \and Francesco Sarnari \thanks{Dipartimento di
Ingegneria dell'Informazione e Scienze Matematiche, Universit\`a di Siena, Via Roma 56, 53100 Italy, e-mail francesco.sarnari@unisi.it}}

\date{}
\maketitle

\vspace{-8mm}
\begin{abstract}

Epilepsy is a neurological disorder arising from anomalies of the electrical activity in the brain, affecting about 0.5--0.8\% of the world population. 
Several studies investigated the relationship between seizures and brainwave synchronization patterns, 
pursuing the possibility of identifying interictal, preictal, ictal and postictal states. 
In this work, we introduce a graph-based model of the brain interactions developed to study synchronization patterns in the electroencephalogram (EEG) signals. The aim is to develop a patient-specific approach, also for a real-time use, for the prediction of epileptic seizures' occurrences.
Different synchronization measures of the EEG signals 
 and easily computable functions able to capture in real-time the variations of EEG synchronization have been considered.
Both standard and ad-hoc classification algorithms have been developed and used.
Results on scalp EEG signals show that this simple and computationally viable processing  
is able to highlight the changes in the synchronization corresponding to the preictal state.
 \\
{\bf Keywords:} EEG analysis; Synchronization measures; Interaction graph; Data classification; Epilepsy.
\end{abstract}

\section{Introduction}\label{sec:intro}

Epilepsy is  a neurological disorder, arising from anomalies of the electrical activity in the brain, affecting about 0.5--0.8\% of the world population.
It represents a very high social cost, resulting in many injuries such as fractures, burns, accidents and even death. 
Treatment options for epilepsy are mainly pharmacological and, to lesser extent, surgical. 
However, antiepileptic drugs have limitations \cite{Deckers} and fail to control seizures in roughly 20--30\% of patients, and surgery is not always possible.
In this context, an important issue is the possibility of predicting epileptic seizure occurrences  (i.e., detecting a pre-ictal or pre-seizure state, if any) in real time, in order to take actions to neutralize an incoming seizure or limit the injuries of a seizure occurrence (e.g., by warning alarms, application of short-acting drugs or electrical stimulation).
The possibility of seizure prediction was explored for over 25 years, typically from the analysis of the electroencephalogram (EEG) signals. 
For recent reviews on this topic, we refer to \cite{rev2017,rev2016, rev2007}. 
Historically, epilepsy has been interpreted as a disorder characterized by abnormally enhanced neuronal excitability and synchronization. 

In this work, a patient-specific graph-based approach is proposed for the prediction epileptic seizure occurrences. The approach is based on the detection of synchronization changes in the electroencephalogram (EEG) signals, which, as stated above, typical occur during the ictal and possibly pre-ictal phase.  Furthermore, the proposed approach has been also designed to be minimally invansive, requiring scalp EEG signals (while most of the studies from the literature are based on intracranial EEG).

Several studies investigated the relationship between seizures and brainwave synchronization patterns, 
highlighting the possibility of distinguishing: {\em interictal}, {\em preictal}, {\em ictal} and {\em postictal} states \cite{Quyen,Mormann2000, Mormann2005,Mirowski}. 
Furthermore, research in the last few years  has replaced the concept of {\em single epileptic focus}  with the concept of  {\em epileptic network} \cite{Wendling, Kramer, Lemieux}. 
Indeed, a network model of the brain interactions appears now more appropriate for the description of epilepsy, 
where the epileptiform activity in any one part is influenced by activity in other parts, 
and the (synchronized) activity of the neurons is involved in the generation of pathological spikes or seizures. 
 
Seizure prediction approaches usually consist of two main phases. In a first phase, a number of measures and indices, generically called features, are computed from the physiological signals 
(typically, from the EEG signals) extracted over time. 
In a second phase, a classification procedure is applied, in order to identify preictal and interictal states \cite{Mormann2005, Parvez} using the time series of the above mentioned features. 
Usually, the ictal and postictal states are discarded from the classification, since the task is to anticipate seizures' occurrences, in order to take suitable actions.

In the first phase, the features can be extracted by using univariate measures, i.e., involving a single EEG channel, or by using multivariate measures, involving two or more EEG channels.
In the literature, many prediction approaches have been based on univariate  measures \cite{Chaovalitwongse,Esteller,Pardalos,Parvez,Rasekhi,Teixeira}. 
However, studies  comparing univariate and bivariate measures \cite{Bandarabadi,Kuhlmann,Mormann2005,Mirowski} 
highlight the good performance in seizure prediction of the features extracted from bivariate measures.


Bivariate measures  naturally lead to a straightforward graph model, taking into account  the scalp morphology  and  the underlying brain interactions. In fact, in our  approach, the nodes of the  graph are associated with the electrodes sites on the scalp and the  weighted edges between nodes  take into the synchronization degree of two EEG signals pairs. 
By using this model, we initially analyze several possible synchronization measures of the EEG signals.  
In particular, the Phase Lock Value (PLV) \cite{Mormann2000}, the Phase Lag Index (PLI) and its weighted version (WPLI) \cite{Vinck} have been tested. 
Subsequently, we develop easily computable functions that should allow us to capture in real-time the variations in the above synchronization.
More specifically, we propose a modified version of a classical indicator called Moving Average Convergence/Divergence (MACD) \cite{Appel} 
commonly used to analyze trends in financial markets.
Finally, we apply both standard classification algorithms to identify the preictal state, namely Support Vector Machines (SVM), and  new ad-hoc linear classifier specifically developed for this application.

DIRE CHE E' PATIENT-SPECIFIC NONINVASIVO E REAL TIME

Data records are obtained by considering the time series generated by the above functions in a rolling time window fashion. 
Computational tests on real data show that the simple and computationally viable processing described above is able to effectively highlight the changes in the synchronization corresponding to the preictal state.

 The paper is organized as follows.
%
 %
Section \ref{sec:ma-me} provides the experimental setup and a block diagram of the whole approach. 
Section \ref{sec:index} presents the graph model of the brain interactions proposed to study the synchronization patterns. 
Section \ref{sec:feat_extract} describes the functions developed to capture the variations in the synchronization patterns. 
Section \ref{sec:selclass} reports our experience in using the classifiers. 
Section \ref{sec:res} provides numerical results on real data from the ``CHB-MIT Scalp EEG Database" \cite{MIT,Shoeb}.
Conclusions follow in Section \ref{sec:conc}.

\section{Materials and methods}\label{sec:ma-me}

\begin{figure}[htbp]
\begin{center}
\includegraphics[width=0.8\textwidth ]{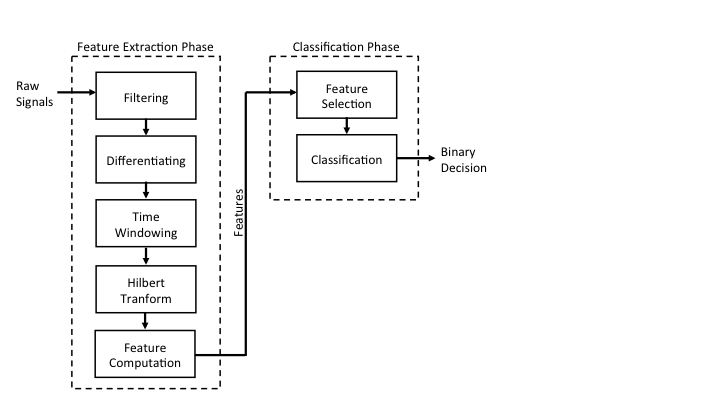}
\caption{Block diagram of the proposed approach}
\label{fig:blockdiagram}
\end{center}
\end{figure}

Seizure prediction methods usually consists of two main phases, each composed of different 
 steps. 
In the first phase (\emph{feature extraction phase}), measures and indices (i.e., features) are computed from physiological signals (typically, from the EEG signals) extracted over time. 
The aim is to transform raw EEG signals given in input into a set of meaningful features that can be used to predict the onset of the crisis. 
In the second phase (\emph{classification phase}), a classification procedure is applied in order to identify preictal and interictal states. 
The aim is to be able to correctly raise an alarm during the preictal period while at the same time to avoid false positive alarms (i.e., triggering an alarm outside the preictal period).
In Figure \ref{fig:blockdiagram} we describe the block diagram of the algorithmic flow of the proposed approach.
The feature extraction phase consists of 5 consecutive steps, while the classification phase is composed of 2 main steps, as explained below.

Steps of the feature extraction phase:
\begin{enumerate}
\item Initially, right after data acquisition, the EEG signal is preprocessed in order to clean it as much as possible from background noise as well as other sources of disturbance, such as artifacts due to eye movements, muscle contractions or even heart beat.
This is obtained by filtering, which is performed by adopting a pass-band filter to each channel of the raw signal. 
The filtering process allows the selection of the band of frequencies of interest, thus removing undesired artifacts. More in details, we adopted a pass-band FIR filter with band $[2,20]$ Hz.

\item Next, we consider the absolute value of the time-derivative of the signal \cite{majumdar}.
This step turns out to be helpful in the analysis of the signal. In fact, differentiating makes the basic noise, nearly flat, even less relevant, while, in contrast, it sharpens the regions where the signal exhibits its peaks, which are most likely to be the regions where seizures occur. 

\item The preprocessing is concluded when the signal is segmented into consecutive time-windows, eventually overlapping, as suggested by different authors \cite{Bandarabadi, Park, Teixeira}. 
Fragmenting the signals into time-windows aims at performing the analysis of the signal in regions with similar and homogenous characteristics in terms of the EEG interpretation.

\item Once the preprocessing is done, we perform, for each channel, the Hilbert transform of the signal, as explained in Section \ref{sec:syncro}. This additional transform allows, for each channel $h$, to construct a complex signal $z_{h}(t)=u_{h}(t)+i\hat{u}_{h}(t)$ with amplitude $A_{h}=\sqrt{u_{h}^{2}+\hat{u}_{h}^{2}}$ and phase $P_{h}=\arctan{(\frac{\hat{u}_{h}}{u_{h}})}$.

\item After that, the desired features are computed from the preprocessed signals. 
In our approach, we use the graph model introduced in Section \ref{sec:graph} with the syncronization measures introduced in Section \ref{sec:syncro}. 
Moreover, we compute additional features, that are the functions developed to capture variations in the synchronization patterns presented in Section \ref{sec:feat_extract}. 
\end{enumerate}

Steps of the classification phase:
\begin{enumerate}
\item In the feature selection step, the set of available features is analyzed to identify the most promising features. Details on the feature selection algorithms are reported in Section \ref{sec:sel}.

\item In the classification step, the features are fed as (training/predicting) input for a binary classifier whose task is to identify the preictal state, thus anticipating  possible seizure onset. 
We use as classifiers Support Vector Machines (see, e.g., \cite{hastie}) and a simpler linear classifier. 
The output of this step is also the output of the whole process and is a binary decision that can be used to alert the patient of the incoming crisis.


\end{enumerate}

\section{A Graph model for the analysis of EEG synchronization}\label{sec:index}
This section is composed of two parts. In the first part (Section \ref{sec:syncro}), different measures of signal synchronization are discussed and evaluated. 
In the second part (Section \ref{sec:graph}), a graph model of the brain interactions is developed to allow the detection of synchronization patterns in EEG signals.

\subsection{Measures of signal synchronization}\label{sec:syncro}

The Phase Lock Value (PLV) or the mean phase coherence \cite{Lachaux, Mormann2000, Mormann2005} 
is one of the most commonly used synchronization measure of EEG signals. 
To compute the phase synchronization, we need to know the instantaneous phase of the two involved signals. This can be extracted using the analytical signal based on the Hilbert Transform, defined as follows: \vspace{-5mm}
$$\hat{x}(t)=\frac{1}{\pi}PV\int_{-\infty}^{\infty}\frac{x(\tau)}{t-\tau}d\tau$$ 
which is well-defined for $x\in L^{p}(\mathbb{R})$, with $1<p<\infty$. 
This additional transform allows the construction of a complex signal $z(t)=x(t)+i\hat{x}(t)=A(t)e^{i\phi(t)}$ with amplitude $A(t)=\sqrt{[ \hat{x}]^2+[ x]^2}$ and phase $\phi(t)=\arctan\frac{\hat{x}(t)}{x(t)}$.

Given channels $h$ and $k$, and a time window $\Delta_t$ containing $N$ instants,  the PLV is defined as follows: 
$$ 
PLV_{h,k,\Delta_t}= \left \vert {1 \over N} \sum_{p=1}^{N}e^{i | \phi_{h}(p)-\phi_{k}(p) | } \right \vert.
$$ 
PLV represents the mean phase coherence of an angular distribution and it takes values in the closed interval $[0,1]$, with a value of 0 corresponding to unsynchronized signals, while 1 to full synchronization. 

More recently, a new measure, called Phase Lag Index ($PLI$), has been introduced in \cite{Stam}. 
This index is based on the idea of discarding the phase differences that center around $0( \mod \pi)$.  
This allows to study  short-term changes of increasing and decreasing synchronization \cite{Winterhalder}. 
In order to discard the phase differences, an asymmetry index is defined by calculating the likelihood that the phase difference $\Delta \phi$ will be in the interval $(-\pi , \pi)$.  
Given the channel pair $h$, $k$ and a time window $\Delta_t$ containing $N$ instants, $PLI$ is given by:
 \begin{equation}\label{PLI}
 PLI_{h,k,\Delta_t} =\left| {1 \over N} \sum_{p=1}^{N} \text{sign} (\phi_{h}(p)-\phi_{k}(p)) \right|
 \end{equation} 

\noindent where  $0\leq PLI \leq 1$. When $PLI=0$  there is either no coupling or a coupling with a phase difference centered around $0(\mod \pi)$, while for $PLI=1$ a perfect phase locking at a value of $\Delta \phi$ different from $0(\mod \pi)$ occurs. The stronger the non zero phase locking is, the larger the $PLI$ will be.

The discontinuity of $PLI$ to small perturbations turns phase lags into leads and vice versa, 
therefore a new measure called Weighted Phase Lag Index ($WPLI$) has been introduced in \cite{Vinck}. 
This index is defined as follows: 
$$WPLI_{h,k,\Delta_t}=\left \vert{1 \over N} \sum_{p=1}^{N}\frac{\vert\sin{(\phi_{h}(p)-\phi_{k}(p))}\vert}{\sin{(\phi_{h}(p)-\phi_{k}(p))}}\right \vert.$$

\noindent
The synchronization indices PLI  and WPLI introduced above are not  computed on the raw EEG signal, but 
on the absolute value of the time-derivative of the signal \cite{majumdar}. 
This approach has been already  successfully used in the literature even for the prediction of epileptic seizures occurrences \cite{Parvez}. 

\begin{figure}[htbp]
\begin{center}
\includegraphics[width=0.6\textwidth ]{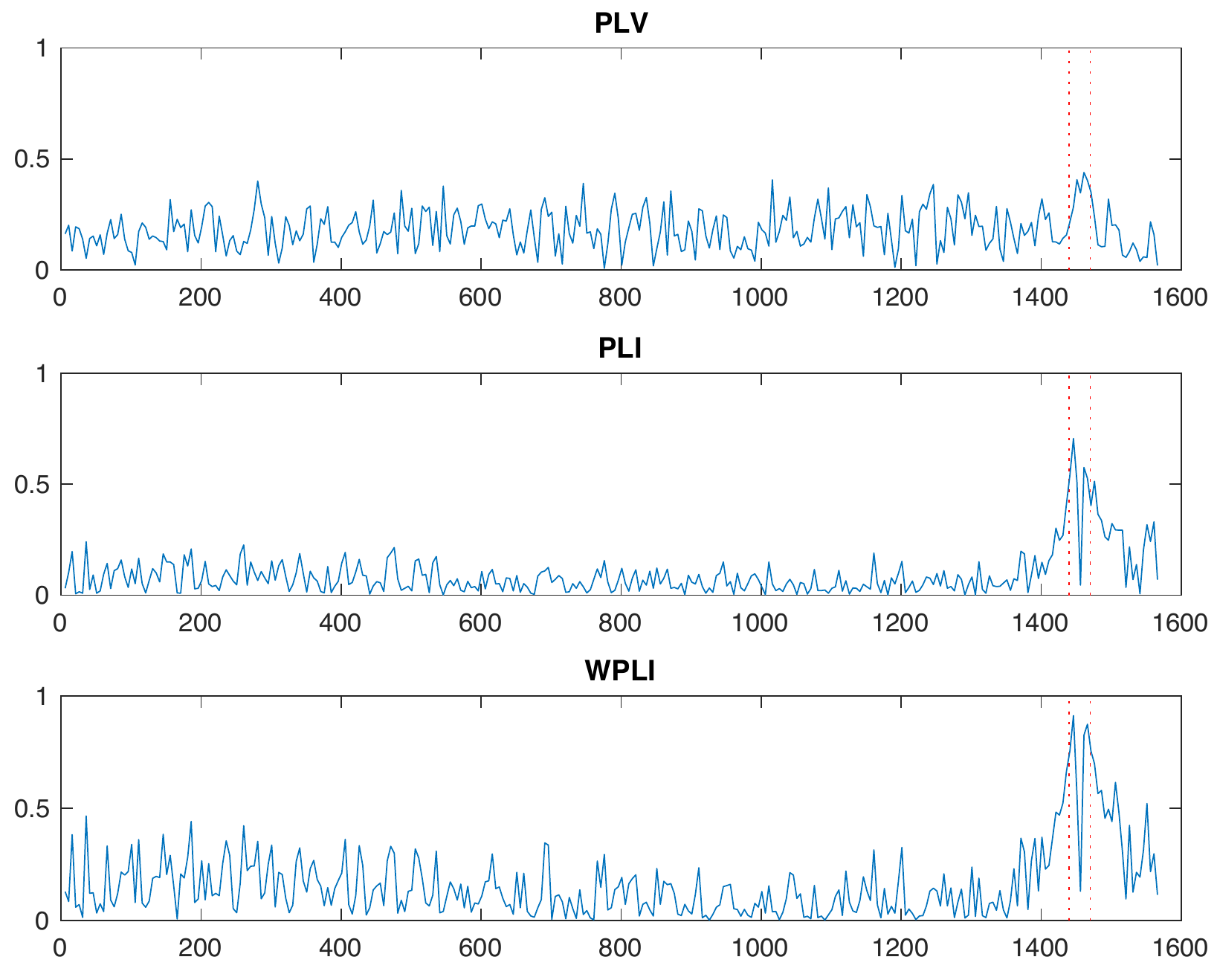}
\caption{PLV, PLI and WPLI on channel pair $\{T7-P7\}$-$\{T7-FT9\}$  of patient $Chb20$ over a time horizon of about 1600 seconds.   }
\label{fig:index1}
\end{center}
\end{figure}

\begin{figure}[htbp]
\begin{center}
\includegraphics[width=0.6\textwidth ]{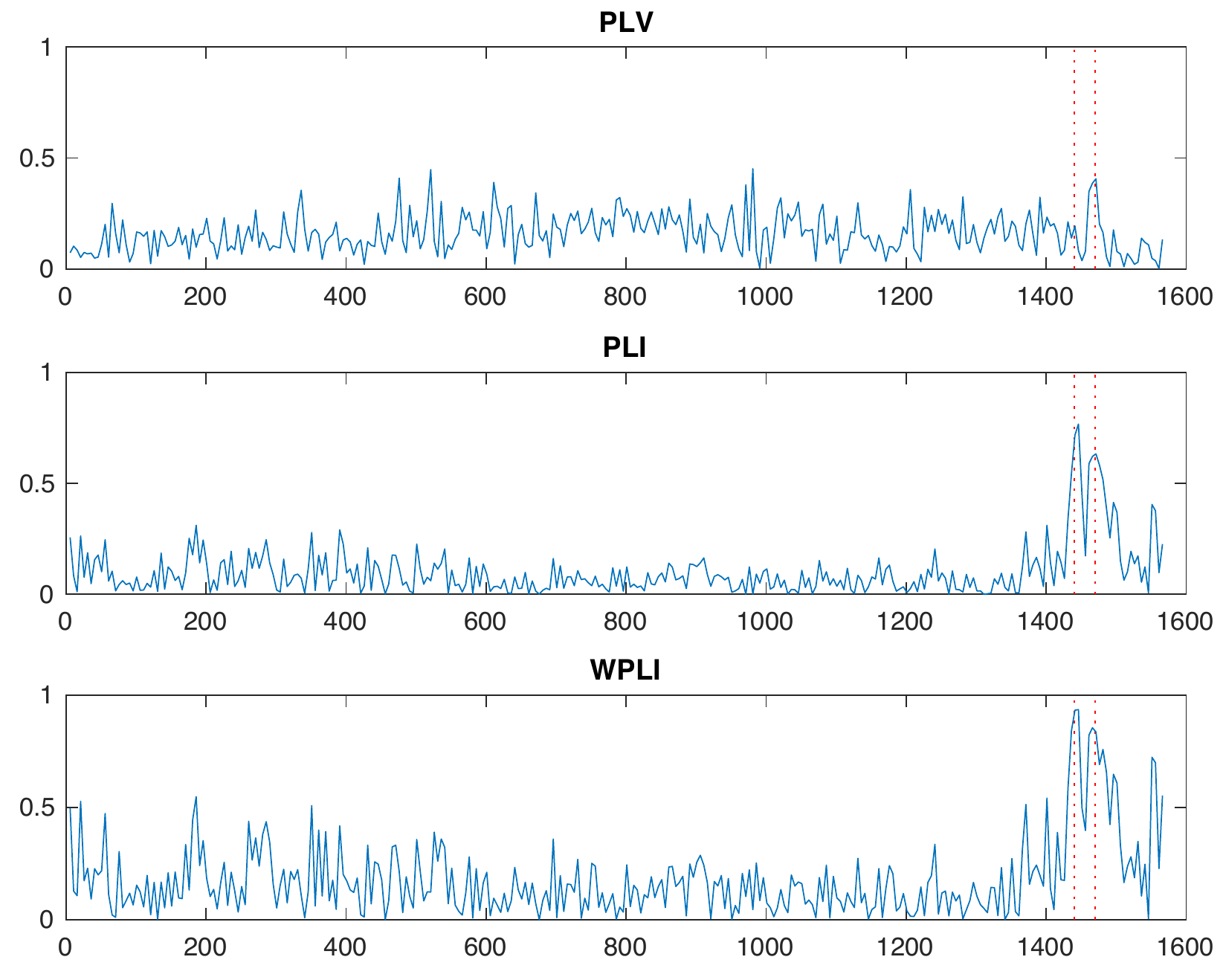}
\caption{PLV, PLI and WPLI on channel pair $\{P4-O2\}$-$\{T7-FT9\}$ of patient $Chb20$  over a time horizon of about 1600 seconds.   }
\label{fig:index2}
\end{center}
\end{figure}

As an example, Figures \ref{fig:index1} and \ref{fig:index2} show the behaviors over time of the functions PLV, PLI and WPLI on two pairs of channels of patient $Chb20$ of  the ``CHB-MIT Scalp EEG Database" \cite{MIT}, computed  using a time window $\Delta_t$ of 6 seconds (with an overlap of 1 sec.). In each graph, the starting and ending times of the epileptic  seizure are marked by the vertical dotted lines.
Observe that PLI and WPLI  have similar behaviors, as they increase as the starting time of the seizure is approaching. 
On the other hand, in both the cases, PLV first decreases  and then increases, but the increase occurs after the beginning of the seizure.

\subsection{Graph model of the brain interactions}\label{sec:graph}

 The synchronization measures introduced in Section \ref{sec:syncro} are symmetric values $w_{h,k}$, defined for each pair of channels $h$ and $k$, i.e., $w_{h,k}=w_{k,h}$.   
In our study,  the raw EEG signals, i.e., the channels, are extracted from  the electrodes positioned on the scalp. 
The connections between  the channels  provide a natural network model  \cite{Rubinov}. 
Such a network can be modeled by an undirected weighted graph $G=(V,A)$, where the nodes represent channels, 
and an undirected weighted edge $(h,k) \in A$ represents the connection between channels $h$ and $k$.  
The weight $w_{h,k}$ associated to the edge $(h,k) \in A$  can be set to the value provided by one of the synchronization measures presented in Section  \ref{sec:syncro}. 
Note that the synchronization measures, and hence the edge weights $w_{h,k}$, vary over time. 
In graph $G$, we assume the presence of an arc only if the measure associated to the pair of channels is greater than a given threshold.


The described  graph model is a mathematical tool which can highlight changes in the neural activity over time, observing how epileptiform events modify  the graph structure. 
Indeed, Graph Theory provides a methodological framework to develop efficient algorithms on the graph for the detection of particular measures and structures, 
which can be used for the analysis of synchronization patterns. 
To this aim, the following graph measures of brain connectivity (see \cite{Rubinov}) could be employed to highlight these patterns:
\begin{itemize}
\item  {\em degree} of a node, i.e.,  the number of edges (with weights larger than a given threshold) connected to a node in $G$;
\item  {\em strength} of a node, i.e.,  the sum of the weights of the edges connected to a node in $G$;
\item {\em clustering coefficient}, i.e.,  is the fraction of triangles around a node, 
which is equivalent to the fraction of node's neighbors that are neighbors of each other.
\item {\em distance and characteristic path length}. The distance is the length of the shortest path between a given pair of nodes. The characteristic path length is the average shortest path length in $G$.
\end{itemize}   

\begin{figure}[hp]
\begin{center}
\includegraphics[width=0.9\textwidth ]{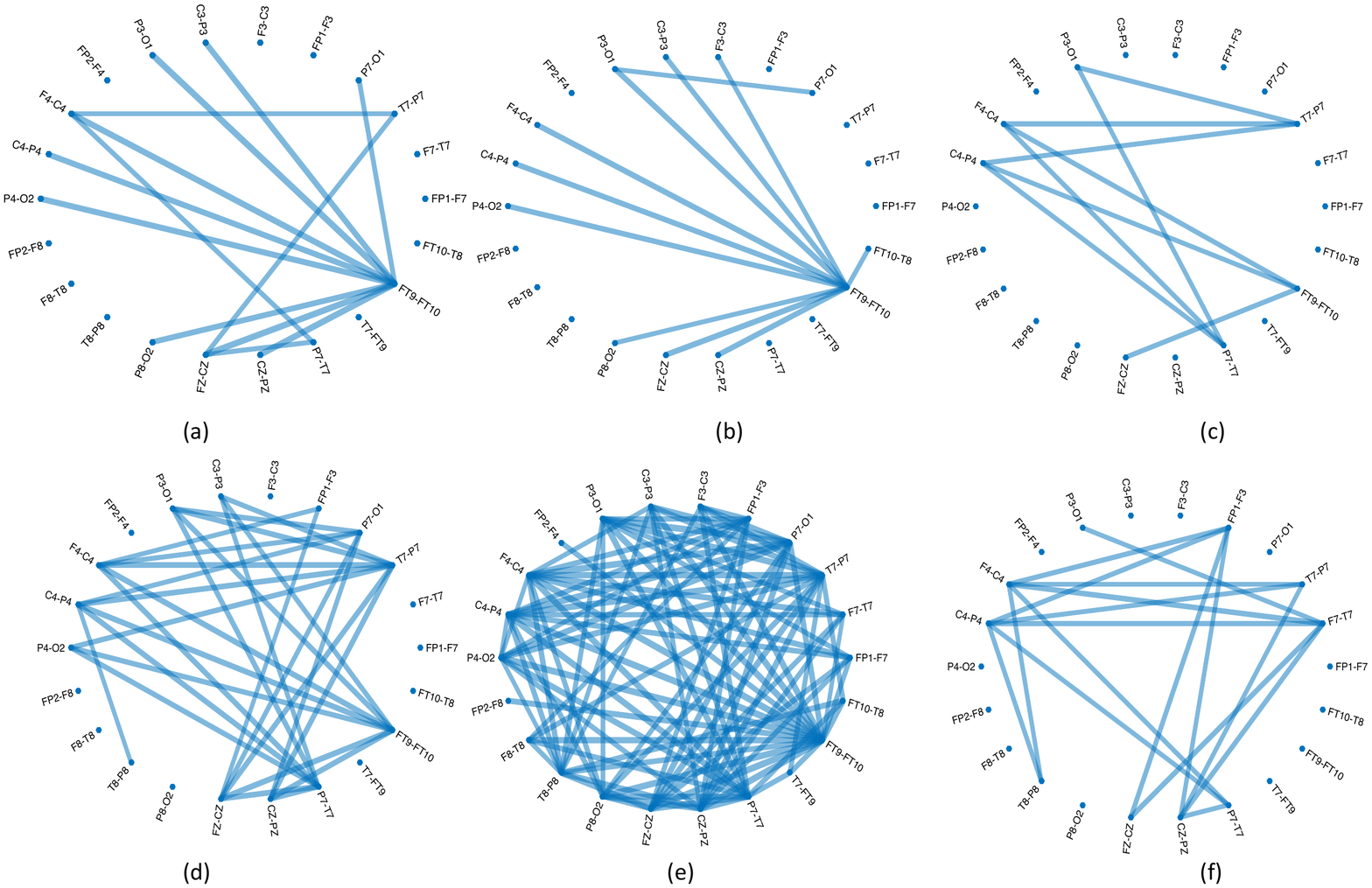}
\caption{(a) $G$  related to   $\Delta_t=[390 ,396]$; (b) $G$  related to   $\Delta_t=[394 ,400]$; (c) $G$  related to   $\Delta_t=[404 ,410]$; (d) $G$  related to   $\Delta_t=[414 ,420]$; (e) $G$  related to  $\Delta_t=[424 ,430]$; (f)  $G$ related to $\Delta_t=[429 ,435]$.   }
\label{fig:graph}
\end{center}
\end{figure}

\noindent
As an example, Figure \ref{fig:graph} illustrates the evolution of the graphs $G$ at several time instants in the approach of an epileptic seizure. 
More precisely, the data are related to the third seizure of the patient $Chb03$ of the ``CHB-MIT Scalp EEG Database" \cite{MIT}, starting at 432 seconds. 
Each graph is related to a time window $\Delta_t$ of 6 seconds (with an overlap of 1 sec.), 
and the edges' weights $w_{h,k}$  measure the Phase Lag Index between channels $h$ and $k$ computed over the time window $\Delta_t$ by Formula \eqref{PLI}. 
Only edges with $w_{h,k}$ larger than 0.7 are reported.
For a better view, the nodes, i.e., the channels, have been positioned around a circle. 
Observe that the edges' weights have a big increase in the two time windows (the graphs in Figures  \ref{fig:graph}.(d) and \ref{fig:graph}.(e)) 
immediately before the seizure (the graph in Figure \ref{fig:graph}.(f)). 

\begin{figure}[!h]
\begin{center}
\includegraphics[width=0.6\textwidth ]{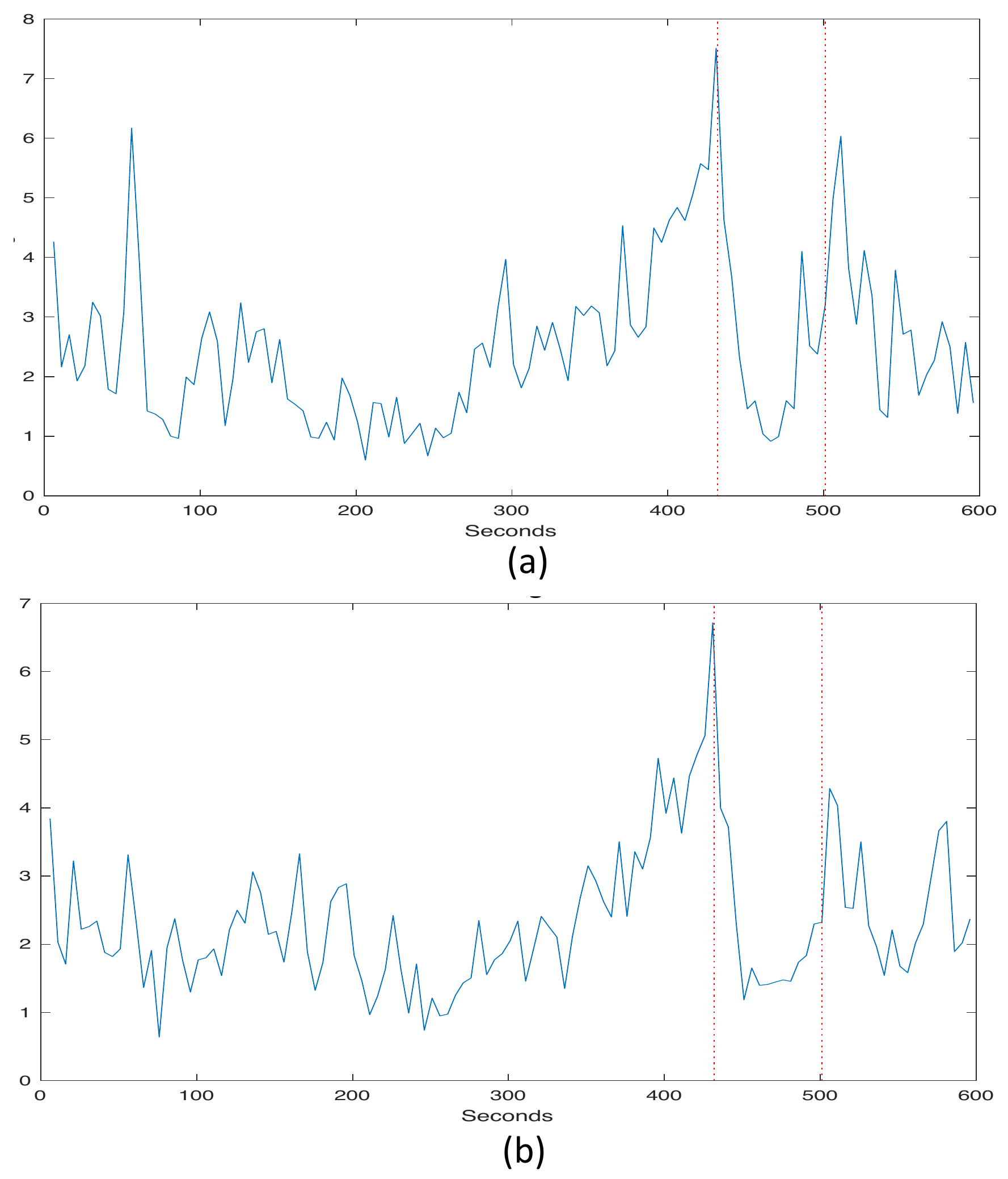}  
\caption{(a) Strength for Channel $T7-P7$; (b) Strength for Channel $P7-O1$.   }
\label{fig:strength}
\end{center}
\end{figure}

Figures \ref{fig:strength}.(a) and \ref{fig:strength}.(b)  report the behavior over time of the node's strength for Channels $T7-P7$ and $P7-O1$, respectivey, on the same data set 
(the third seizure of patient $Chb03$, starting at second 432 seconds). The two vertical dotted lines delimit the ictal period.
Note that the strengths of the nodes increase as the ictal period is approaching and then sharply decrease. The period immediately preceding the ictal period is the preictal period. In our experiments, we consider  the spanning of the preictal period as a parameter and we call it {\em prediction interval}. 

\section{Highlighting variations in the EEG synchronization}\label{sec:feat_extract}

The prediction of a seizure from the analysis of the synchronization measures introduced in Section \ref{sec:index} could already be viewed as a classification problem. 
However, to successfully use a classification algorithm, we should integrate the features directly extracted from the EEG data, i.e.,  
the above synchronization measures,  with additional information derived from them.
This additional information should basically inform the classifier, in each time instant, about the relation occurring at that time between 
the current value of the features and their past values considered at an aggregate level.
Indeed, current data alone do not contain enough information to allow a reliable detection of a preictal  state, 
and we need to define functions that should be able to highlight the particular rising trends in the synchronization measures
which are precursors of a seizure occurrence (see also Figure \ref{fig:strength}). 

Similar trend analysis problems have been studied in depth in the field of Finance.
Forecasting trends in the value of an asset on a financial market is indeed one very basic issue in this field. 
Of course, such values constitute stochastic processes, and any similar prediction attempt has to deal with uncertainty.
However, the {\it technical analysis of prices} is an analysis methodology precisely developed to forecast the direction of the prices of a security
through the study of past market data. This is done by computing a number of indicator functions whose value should support in the prediction task.
More details can be found for instance in \cite{Kirkpatrick,Colby}.
Though the practical effectiveness of these methodologies can be questioned, since in real-world the market evolution is deeply influenced also by external factors,
the indicator functions developed in this field may be of help in the detection of the above described  trends in the features extracted from the EEG data.
In particular, by denoting each of the feature computed in the previous section by  $f_h(t)$, with $ h  \in \{ 1,\dots,H \}$, 
we resort to the following functions:
\begin{enumerate}

\item A function $T_h (t)$ describing the {\it trend} of the generic feature $f_h (t)$ 
at time period $t$, with $ h  \in \{ 1,\dots,H \}$.

\item A function $L_h (t)$ representing a {\it current lower limit} of the above $T_h (t)$ 
in a time window representing the recent past with respect to time period $t$, with $ h  \in \{ 1,\dots,H \}$.

\item A function $M_h (t)$  measuring the {\it elevation} of the current trend above the current lower limit at time period $t$, 
in order to detect whether a rising trend occurs for a sufficient interval of time, with $ h  \in \{ 1,\dots,H \}$.

\end{enumerate}

\noindent
\paragraph{Choice of the trend function $T_h (t)$.}
A moving average, i.e., the average over a certain time interval of the values of $f_h(t)$, is generally chosen to describe trends. 
A weighted moving average (WMA) has multiplying factors to give different weights to the different instants of time. 
Usually, recent instants receive more importance than older ones. 
In particular, an exponential moving average (EMA) applies weighting factors which decrease exponentially in the past, using a parameter $w$ representing the extension of the past.  
In our case, we chose for the trend function $T_h (t)$ an EMA, computed as follows. 
$$
T_h (t) = \left \{ \begin{array}{ll} 
f_h (1) & {\rm for \ \ } t=1 \\ 
 \left(\frac{2}{w+1}\right) f_h (t) +  \left(1- \frac{2}{w+1} \right) T_h (t-1) & {\rm for \ \ } t>1 
\end{array} \right.
$$
We experimentally find that $w$ = 7 provides a good trend description in our application. 

\bigskip
\noindent
\paragraph{Choice of the current lower limit $L_h (t)$.} To evaluate the current lower limit, we chose the minimum of $T_h (t)$ over the previous $p$ time intervals.
$$
L_h (t)= \min_{\tau \in \{t-p,\dots, t\} }\{T_h (\tau)\}
$$
We experimentally find that $p$ = 27 provides a satisfactory lower limit in our application. 

\bigskip
\noindent
\paragraph{Choice of the elevation function $M_h (t)$.}  In this case, we follow the ideas underlying the trading indicator called Moving Average Convergence/Divergence (MACD). 
This indicator should reveal changes in the strength, direction, and momentum of a trend in the  price of an asset \cite{Appel}. 
The simplest version of MACD is the difference between two moving averages, one over a shorter time interval and one over a longer time interval.
Indeed, when the trend is increasing, the moving average over the shorter interval becomes the greater one.
Conversely, the same moving average becomes the smaller one when the trend is decreasing.
Further insight can be obtained by using a third moving average of the MACD  itself over an even shorter interval, called "signal line". 

However, to detect the particular kind of rising trend that in our application represents a preictal situation, 
we experienced better accuracy by substituting 
the longer-period average with the current lower limit $L_h (t)$. 
This allows to highlight not only the  ``relative" information of the changes in the trend of a feature $f_h (t)$, 
but also the more  ``absolute" information of the amplitude of the elevation of that feature over the current low value.
We call this difference Moving Average and Amplitude Convergence/Divergence (MAACD), computed as follows:
\begin{equation}\label{MAACD}
M_h (t) = T_h (t) - L_h (t) \ \ \ \ \ \ \  h  \in \{ 1,\dots,H \}.
\end{equation}



\noindent
In conclusion, we compute the described MAACD for each feature $f_h$, with $ h  \in \{ 1,\dots,H \}$, and we add it to the set of the features available to perform the classification.

\section{Feature selection and classification}\label{sec:selclass}
This section describes both the feature selection and the classification approaches used to predict the seizures.

\subsection{Feature selection}\label{sec:sel}
An important step in the classification task is the feature selection phase. 
Basically, the selection phase uses a variety of search techniques for identifying the subsets of features that are the most {\em relevant} for the classification task,
possibly providing a measure which scores the different subsets. 

Two different feature selection approaches have been considered and evaluated. 
The first approach  has been specifically developed for the problem under study. 
It is a threshold-based approach and is composed of two steps. 
In the first step, a set of thresholds, one for each available feature, is obtained from the training set. 
In practice, threshold $th_h$ corresponding to feature $f_h$ is computed as the average over time of $f_h$ in the training set.
In the second step, two different rankings of the features are computed. 
The first ranking is obtained by counting the number of time periods for which $f_h$ has a value above its threshold $th_h$ 
inside the time intervals used for prediction, corresponding to preictal states. This ranking basically evaluates, for each feature $f_h$, its ability in the identification of positive records, i.e.,  those corresponding to preictal states.
The second ranking is given by the number of time periods for which $f_h$ has a value below its threshold $th_h$ outside the prediction intervals. 
Hence, it basically evaluates the ability in not producing false positive predictions of the negative records, i.e., those corresponding to interictal states.
The length of the time intervals used for prediction is an algorithmic parameter and will be detailed in Section \ref{sec:res}.

The second approach uses the standard feature selection technique called Relief  \cite{Kira}. 
Although several variants of the Relief algorithm have been proposed in the literature, the main idea of the original Relief algorithm  
is  to evaluate the quality of each feature according  to 
its capability of distinguishing between similar records of opposite classes more than it does between similar records of the same class.
This operation is performed using records of the training set.

\subsection{Classification}\label{sec:class}
In classification, the objective is to identify the classes new records belong to. 
Given a set of records grouped into classes, i.e., labeled, the
classification task consists
in learning from them a criterion to assign the class to new unlabeled
records \cite{hastie,slad}.
Two classification approaches have been considered and evaluated for our
problem.
The first approach consists in the use of a linear classifier specifically designed for this problem, while the second one is the use of a Support Vector Machine (SVM) algorithm from the literature. 
In both approaches, the classifier works using the features selected in the feature selection phase. 

In the first approach, a subset of features is selected according to the rankings obtained by the threshold-based algorithm for feature selection.
Then, the features $f_h$ and their corresponding thresholds $th_h$ (computed as in Section \ref{sec:sel}) are linearly combined to obtain a single feature and threshold. 
In this case, the classification is performed by evaluating, on the test data set, the number of times the combined feature is above (below) its threshold inside (outside) the time intervals used for prediction, corresponding to preictal states.

In the second classification  approach, we use the  Least Square-SVM (LS-SVM) algorithm \cite{Suykens}.  
While in standard SVMs the solution of the classification problem is obtained by a convex quadratic programming problem, in LS-SVM  a least squares cost function is employed, so as to obtain a set of linear equations in the dual space. 
Such a choice allows to reduces the computational burden of the solution of the constrained optimization problem. LS-SVM has been recently and successfully used for seizure prediction in \cite{Parvez}.

\section{Data description and computational analysis}\label{sec:res}

This
section is organized as follows.
In Section \ref{sec:dd} we introduce and describe the EEG recordings used in our experiments. Next, in Section \ref{sec:ad} we report the details of the compared algorithms.
Then, we first provide aggregated results (Section \ref{sec:ar}), and finally, in Section \ref{sec:dr}, we report the detailed results for all the patients considered.

\subsection{Data description}\label{sec:dd}

We considered 10 patients from the ``CHB-MIT Scalp EEG Database'' \cite{MIT,Shoeb}, which consist of scalp EEG recordings from pediatric subjects with intractable seizures from the Children's Hospital Boston.
All signals are sampled at 256 Hz using the International 10--20 system of EEG electrode positions.
EEG signals are filtered using a band-pass FIR filter with band [2--20] Hz.

Table \ref{pat_info} shows the patients considered in our analysis, and, for each of them, the number of EEG channels and of seizures used. 
For the analysis, we have selected patients having a suitable number of seizures and with seizures sufficiently sparse over time (hopefully containing both interictal and preictal states).  
 For each patient and seizure, a data set has been extracted from the raw EEG, ending with the beginning of the seizure and starting at most 3600 seconds before it.
When the seizure starts earlier than 3600 seconds in the raw EEG data, all the data records until the seizure have been selected. 
As a consequence, 53 datasets have been considered (1 for each patient/seizure pair). 

{\small
\begin{table}
\begin{center}
\begin{tabular}{l r r r r}
Pat. id &\# Channels &\# Seizures\\
\hline
Chb01&22&7\\
Chb03&22&7\\
Chb05&22&5\\
Chb08&22&5\\
Chb15&23&5\\
Chb18&22&5\\
Chb20&22&5\\
Chb21&22&4\\
Chb23 &22&5\\
Chb24 &22&5\\
\hline
\end{tabular}
\end{center}
\caption{Patients, channels and seizures of the CHB-MIT Scalp EEG Database analyzed.}
\label{pat_info}
\end{table}
}

In the feature extraction phase, the following steps have been performed on each data set:  $(i)$ The synchronization measures presented in Section \ref{sec:syncro} have been computed for each channel pair on a time window of 6 secs with an overlap of 1 sec; $(ii)$ the node strength has been computed for each channel (as defined at the end of Section \ref{sec:graph}); $(iii)$ the MAACD feature are computed on each node strength, as defined in  \eqref{MAACD}.

For each patient of Table \ref{pat_info},  $n_t=\lceil \# Seizures /2 \rceil$ data sets (each containing a single seizure) have been used for the training phase,  and the remaining $\# Seizures-n_t$ data sets for the test phase.

\subsection{Details of the classification algorithms}\label{sec:ad}

As described in Section \ref{sec:selclass}, two classification algorithms have been developed. One, called $R-SVM$, in which the Relief algorithm and the LS-SVM algorithm have been used for the feature selection and classification phases, respectively. The other, called $TH$, in which both the selection and the classification  phase are performed by a threshold-based approach (described in Section \ref{sec:selclass}). 
All the algorithms have been coded in Matlab, the Fieldtrip toolbox \cite{fieldtrip} has been used for EEG data acquisition and filtering. 
In the SVM of algorithm $R-SVM$,  RBF kernel is used, since it has been shown in  \cite{Parvez} to have the best  performances for this kind of classification problems.
The regularization parameter $\gamma$, determining the trade-off between the training error minimization and smoothness, and  the squared bandwidth $\sigma^2$ are found by a tuning step performed with the simplex method.

The $R-SVM$ and $TH$ algorithms have been tested with different parameter configurations and subsets of features. 
For both $R-SVM$ and $TH$, in the training phase, we set to True the preictal period ranging from $T$ seconds before the seizure onset till the actual onset of the seizure of  each dataset (i.e., $T$ is the prediction interval). 
All the other periods instead are set to False. The same  prediction interval $T$ has been used in the test phase.
In the computational experiments, three different lengths of the prediction interval $T$ have been evaluated, i.e., $T \in \{150, 200, 300\}$ seconds.

In the results of $R-SVM$, 
we denote by $np$ the length of the sliding window (containing $np$ consecutive points) used to build  the classes of each dataset, and by $Feat.$ the subset of features used in the classification phase. 
$R-SVM$ has been tested with $np\in \{2,5,10\}$, and $Feat.$ chosen in one of the following ways:
\begin{itemize}

\item $Str.R1$: the feature selection phase is performed by Relief only considering  the strengths of the nodes.
The LS-SVM classifier uses the strength of the first node in the ranking provided by Relief.

\item$Str.LC$: the feature selection phase is performed by Relief  only considering  the strengths of the nodes. 
The LS-SVM classifier uses a feature obtained by linearly combining the strengths of the first four nodes in the ranking (provided by Relief), weighted by the related weights provided by Relief.

\item $MAACD.R1$: the feature selection phase is performed by Relief only considering  the MAACD features, computed on the strengths of the nodes. 
The LS-SVM classifier uses the first MAACD feature in the ranking provided by Relief.

\item$MAACD.LC$: the feature selection phase is performed by Relief only considering  the MAACD features, computed on the strengths of the nodes. 
The LS-SVM classifier uses a feature obtained by linearly combining the first four  MAACD features in the ranking provided by Relief, weighted by the related weights provided by Relief.

\end{itemize}
In algorithm $TH$,  the feature selection phase is performed by the threshold-based approach considering only the MAACD features, computed on the strengths of the nodes.
Algorithm $TH$  depends on the three parameters: $a_1$, $a_2$ and $a_{TH}$.  
Parameters $a_1$ and $a_2$ are two coefficients used to build the feature employed in the classification phase. 
More precisely, in the feature selection phase, two rankings are built by using the MAACD features. 
As stated in Section \ref{sec:selclass}, in the first ranking, the MAACD functions are ordered, in ascending order, according to the number of times each $MAACD_h$ feature is above its threshold $th_h$ outside the  prediction interval $T$. 
Observe that the first MAACD feature in this ranking provides the smallest number of false positives. 
In the second ranking, the MAACD functions are ordered, in descending order, according to the number of times each $MAACD_h$ feature is above its threshold inside  the prediction interval $T$. 
(The first MAACD feature in this ranking provides the greatest number of true positives.) 
Letting $MAACD_h$ and $MAACD_k$ be the first two features in the two rankings, the feature used in the classification phase is given by
$$a_1 MAACD_h + a_2 MAACD_k.$$ 
Parameter $a_{TH}$ is used to build the  threshold used in the classification phase, as follows. Letting $th_h$ and $th_k$ be the thresholds related to $MAACD_h$ and $MAACD_k$, respectively, computed in the feature selection phase, the  threshold used in the classification phase is given by 
$$a_{TH}(a_1 th_h + a_2 th_k).$$ 

\noindent
Algorithm $TH$ has been tested with the following values of parameters:\\ $(a_1,a_2) \in \{(1,0);(0,1);(0.5,0.5);(0.25,0.75);(0.75,0.25)\}$; $a_{TH} \in \{1.1,1.25,1.5,1.75,2,2.25\}$.


\subsection{Aggregated results}\label{sec:ar}

Table \ref{aggregateresults} provides the aggregated results obtained by the two algorithms $R-SVM$ and $TH$, 
considering for each patient the best performances of each algorithm. 
In particular, a performance of an algorithm has been classified better than another, if it provides a smaller number of false positives, is able to predict a larger number of seizures and, in a second analysis, yields a larger number of true positives. 

Each row represents one of the patients  from the "CHB-MIT Scalp EEG Database" described  in Table  \ref{pat_info}. 
The first column of the table show the patient's id. The next four columns report the FP, TP, Miss and $\Delta$ for the $R-SVM$ approach, whereas the last four columns report the same data for the threshold-based method ($TH$). More specifically, for each algorithm, FP and TP are the number of false and true positives, respectively. Miss is the number of missed seizures (i.e., the  seizures not detected by the algorithm). While $\Delta$ is the prediction time, computed as the average time (in seconds) from the seizure onset in which the first true positive occurs. In the case in which a seizure is not predicted by an algorithm, then its contribution to $\Delta$ is set to 0.

{\small
\begin{table}
\begin{center}
\begin{tabular}{c|r r r r|r r r r}
        & \multicolumn{4}{c|}{$R-SVM$} & \multicolumn{4}{c}{$TH$} \\
Patient & FP & TP & Miss & $\Delta$ & FP & TP & Miss & $\Delta$\\
\hline
$Chb01$ &  1 &  9 & 0 &  6.0 &  0 & 54 & 0 &  87.6 \\
$Chb03$ &  0 & 19 & 0 & 44.0 &  0 & 19 & 0 & 44.0 \\
$Chb05$ &  0 & 10 & 0 & 31.5 &  0 & 26 & 1 & 110.0 \\
$Chb08$ & 10 &  1 & 1 & 84.5 &  0 &  8 & 1 &  17.0 \\
$Chb15$ &  0 & 14 & 0 & 29.5 &  0 & 12 & 0 &  24.5 \\
$Chb18$ &  0 & 32 & 0 & 71.5 &  0 & 30 & 0 &  66.0 \\
$Chb20$ &  0 &  3 & 1 &  4.0 &  0 & 33 & 0 &  76.5 \\
$Chb21$ &  0 &  2 & 1 &  1.0 & 0 & 14 & 0 & 33.5 \\
$Chb23$ &  0 &  7 & 1 & 34.5 &  0 &  2 & 1 &   7.0\\
$Chb24$ &  0 & 18 & 0 & 36.5 &  0 & 17 & 0 &  34.0 \\
\hline
Avg.   & 1.1&11.5&0.4&34.3&0&21.5&0.3&50.0\\
\hline
\end{tabular}
\end{center}
\caption{Aggregate results and algorithms' comparison.}
\label{aggregateresults}
\end{table}
}

From Table \ref{aggregateresults} we observe as $TH$ performs sightly better than $R-SVM$.
In fact, $TH$  produces a lower number of false positives (equal to 0) than $R-SVM$. 
When considering the true positives (i.e., the number of times the algorithm correctly predicts the approaching  of a seizure), $TH$ performs better than  $R-SVM$, too.
The two algorithms have comparable performance in term of number of missed seizures. Observe that,  on all the considered patients  and seizures, $R-SVM$ and $TH$ miss 4 and 3 occurrences over 21, respectively.
Finally, when considering the prediction value $\Delta$, we observe that $TH$ is again able to predict the onset of a seizure with a sightly greater advance than the $R-SVM$. 
This could be particularly valuable in order to take suitable actions able to neutralize the seizure or limit its consequences. 

\subsection{Detailed results}\label{sec:dr}

In Tables \ref{Chb01}--\ref{Chb24}, detailed results for the two classification algorithms on patients of Table \ref{pat_info} are given. In each table, the best three parameter configurations (i.e., those providing the best results) of each algorithm are reported, where $np$ and  $Feat.$ are the parameters of $R-SVM$, and $a_1$, $a_2$ and $a_{TH}$ are  parameters of $TH$, while $T$ is the prediction interval. All these parameters have been described in Section \ref{sec:ad}.
As already stated in the previous section, the performance of an algorithm has been considered better than another, if 
it provides a smaller number of false positives, is able to predict a larger number of seizures and, in a second analysis, yields a larger number of true positives. 

In these tables, for both algorithms, FP and TP are the number of false  and true positives, respectively, \newline 
\#  seiz. is "Yes" if all seizures of the test set are predicted on $T$ (i.e., if a TP exists in each prediction interval of each patient test set), 
and "Not" otherwise,  $\Delta_i$ reports on the seconds from each seizure onset of the test set in which the first true positive occurs, for $i \in \{1,\ldots,\# Seizures-n_t\}$.

The detailed analysis shows that algorithm $R-SVM$ is always able to detect all the seizures of patients $Chb03$, $Chb15$, $Chb18$ and $Chb24$,
while algorithm $TH$ all those of patients $Chb01$, $Chb18$, $Chb20$  and $Chb24$. 
In terms of false positives, $TH$ appears more robust than $R-SVM$ providing no false positive, while false positives are found by $R-SVM$ on 5 patients. 
Regarding the features employed by $R-SVM$, the MAACD features provide the best results 18 times out of 30, while the node strength 13 times out of 30.
For algorithm $TH$, there is not a setting of parameters $a_1$ and $a_2$ performing  better, while all the best results are obtained with $a_{th}\geq 1.5$, except in one case (see Table \ref{Chb21}).


{\small
\begin{table}
\begin{center}
\begin{tabular}{l r r r r r r r r r r r r}
Algo&&$np$& $Feat.$ & $T$& FP& TP & \# seiz.& $\Delta_1$& $\Delta_2$& $\Delta_3$\\
\hline$R-SVM$&&10&$Str.LC$&150&1&9&Yes&9&6&3\\
$R-SVM$&&2&$MAACD.R1$ &300&1&5&Yes&0&1&0\\
$R-SVM$&&5&$MAACD.R1$ &300&1&1&Not&139&  -&-\\
\hline\hline
Algo&$a_1$&$a_2$&$a_{TH}$&$T$& FP & TP  & \# seiz. & $\Delta_1$ & $\Delta_2$ & $\Delta_3$ \\
\hline$TH$&0&1&2&300&0&54&Yes &99&61&103\\
$TH$&0.25 &0.75 &1.75 &300&0&52&Yes &94&61&103\\
$TH$&1&0&1.5 &150&0&47&Yes &89&61&78\\
\hline
\end{tabular}
\end{center}
\caption{Results on patient $Chb01$.}
\label{Chb01}
\end{table}
}

{\small
\begin{table}
\begin{center}
\begin{tabular}{l r r r r r r r r r r r r}
Algo&&$np$& $Feat.$ & $T$& FP& TP & \# seiz.& $\Delta_1$& $\Delta_2$& $\Delta_3$\\
\hline$R-SVM$&&2&$MAACD.R1$ &200&0&19&Yes&21&93&18\\
$R-SVM$&&2&$Str.R1$&200&0&13&Yes&6&13&13\\
$R-SVM$&&5&$MAACD.LC$ &200&0&11&Yes&1&13&8\\
\hline\hline
Algo&$a_1$&$a_2$&$a_{TH}$&$T$& FP & TP  & \# seiz. & $\Delta_1$ & $\Delta_2$ & $\Delta_3$ \\ \hline
$TH$&0.25 &0.75 &2.25 &300&0&19&Yes &6&93&33\\
$TH$&0&1&1.5 &150&0&12&Yes &11&8&8\\
$TH$&0.5 &0.5 &2.25 &300&0&20&Not &93&143&-\\
\hline
\end{tabular}
\end{center}
\caption{Results on patient $Chb03$.}
\label{Chb03}
\end{table}
}

{\small
\begin{table}
\begin{center}
\begin{tabular}{l r r r r r r r r r r r }
Algo&&$np$& $Feat.$ & $T$& FP& TP & \# seiz.& $\Delta_1$& $\Delta_2$\\
\hline$R-SVM$&&10&$Str.LC$&150&0&10&Yes&50&13\\
$R-SVM$&&5&$Str.R1$&150&2&1&Not&5&-\\
$R-SVM$&&5&$Str.LC$&150&3&8&Yes&45&13\\
\hline\hline
Algo&$a_1$&$a_2$&$a_{TH}$&$T$& FP & TP  & \# seiz. & $\Delta_1$ & $\Delta_2$ \\ \hline
$TH$&1&0&1.75 &300&0&26&Not &220&-\\
$TH$&0&1&1.75 &300&0&26&Not &220&-\\
$TH$&0.5 &0.5 &1.75 &300&0&26&Not &220&-\\

\hline
\end{tabular}
\end{center}
\caption{Results on patient $Chb05$.}
\label{Chb05}
\end{table}
}

{\small
\begin{table}
\begin{center}
\begin{tabular}{l r r r r r r r r r r r }
Algo&&$np$& $Feat.$ & $T$& FP& TP & \# seiz.& $\Delta_1$& $\Delta_2$\\
\hline$R-SVM$&&10&$Str.R1$&300&10&1&Not &169&-\\
$R-SVM$&&2&$Str.R1$&150&13&3&Not &29&-\\
$R-SVM$&&2&$MAACD.LC$ &150&17&6&Not &24&-\\
\hline\hline
Algo&$a_1$&$a_2$&$a_{TH}$&$T$& FP & TP  & \# seiz. & $\Delta_1$ & $\Delta_2$ \\ \hline
$TH$&0.5 &0.5 &1.75 &200&0&8&Not &34&-\\
$TH$&0.5 &0.5 &1.75 &150&0&4&Not &29&-\\
$TH$&0.5 &0.5 &2&150&0&2&Not &4&-\\
\hline
\end{tabular}
\end{center}
\caption{Results on patient $Chb08$.}
\label{Chb08}
\end{table}
}

{\small
\begin{table}
\begin{center}
\begin{tabular}{l r r r r r r r r r r r }
Algo&&$np$ & $Feat.$  & $T$ & FP & TP  & \# seiz. & $\Delta_1$ & $\Delta_2$ \\ \hline
$R-SVM$&&2&$MAACD.R1$  &200&0&14&Yes &50&9\\
$R-SVM$&&2&$MAACD.LC$  &200&0&14&Yes &50&9\\
$R-SVM$&&2&$MAACD.R1$  &150&0&12&Yes &45&4\\
\hline\hline
Algo&$a_1$&$a_2$&$a_{TH}$&$T$& FP & TP  & \# seiz. & $\Delta_1$ & $\Delta_2$ \\ \hline
$TH$&0&1&1.5 &200&0&12&Yes &45&4\\
$TH$&1&0&1.5 &300&0&12&Not &60&-\\
$TH$&0.75 &0.25 &1.5 &300&0&10&Not &45&-  \\
\hline
\end{tabular}
\end{center}
\caption{Results on patient $Chb15$.}
\label{Chb15}
\end{table}
}

{\small
\begin{table}
\begin{center}
\begin{tabular}{l r r r r r r r r r r r }
Algo&&$np$ & $Feat.$  & $T$ & FP & TP  & \# seiz. & $\Delta_1$ & $\Delta_2$ \\ \hline
$R-SVM$&&10&$MAACD.R1$  &200&0&32&Yes &87&55\\
$R-SVM$&&10&$MAACD.LC$  &200&0&31&Yes &82&55\\
$R-SVM$&&2&$MAACD.R1$  &150&0&30&Yes &77&55\\
\hline\hline
Algo&$a_1$&$a_2$&$a_{TH}$&$T$& FP & TP  & \# seiz. & $\Delta_1$ & $\Delta_2$ \\ \hline
$TH$&0&1&1.5 &300&0&30&Yes &77&55\\
$TH$&1&0&1.5 &150&0&26&Yes &72&45\\
$TH$&0.25 &0.75 &1.5 &300&0&26&Yes &72&45\\
\hline
\end{tabular}
\end{center}
\caption{Results on patient $Chb18$.}
\label{Chb18}
\end{table}
}

{\small
\begin{table}
\begin{center}
\begin{tabular}{l r r r r r r r r r r r }
Algo&&$np$ & $Feat.$  & $T$ & FP & TP  & \# seiz. & $\Delta_1$ & $\Delta_2$ \\ \hline
$R-SVM$&&2&$MAACD.R1$  &150&0&3&Not &8&-\\
$R-SVM$&&5&$MAACD.R1$  &150&0&3&Not &13&-\\
$R-SVM$&&5&$MAACD.R1$  &200&0&3&Not &13&-\\
\hline\hline
Algo&$a_1$&$a_2$&$a_{TH}$&$T$& FP & TP  & \# seiz. & $\Delta_1$ & $\Delta_2$ \\ \hline
$TH$&1&0&1.5 &300&0&33&Yes &90&63\\
$TH$&0.75 &0.25 &1.5 &300&0&31&Yes &80&63\\
$TH$&1&0&1.5 &200&0&29&Yes &80&63\\
\hline
\end{tabular}
\end{center}
\caption{Results on patient $Chb20$.}
\label{Chb20}
\end{table}
}

{\small
\begin{table}
\begin{center}
\begin{tabular}{l r r r r r r r r r r r r}
Algo&&$np$ & $Feat.$  & $T$ & FP & TP  & \# seiz. & $\Delta_1$& $\Delta_2$ \\ \hline
$R-SVM$&&2&$MAACD.R1$&150&0&2&Not&2&-\\
$R-SVM$&&2&$Str.R1$&150&2&2&Yes &2&-\\
$R-SVM$&&2&$Str.R1$&200&2&2&Yes &2&-\\
\hline\hline
Algo&$a_1$&$a_2$&$a_{TH}$&$T$& FP & TP  & \# seiz. & $\Delta_1$ & $\Delta_2$ \\ \hline
$TH$&1&0&1.25 &150&0&14&Yes&62&5\\
$TH$&1&0&1.5 &150&0&7&Not&52&-\\
$TH$&0.75 &0.25 &1.5 &150&0&4&Not&47&-\\
\hline
\end{tabular}
\end{center}
\caption{Results on patient $Chb21$.}
\label{Chb21}
\end{table}
}

{\small
\begin{table}
\begin{center}
\begin{tabular}{l r r r r r r r r r r r r}
Algo&&$np$ & $Feat.$  & $T$ & FP & TP  & \# seiz. & $\Delta_1$ & $\Delta_2$ \\ \hline
$R-SVM$&&5&$Str.R1$ &150&0&7&Not &69&-\\
$R-SVM$&&10&$Str.R1$ &200&1&6&Yes &54&4\\
$R-SVM$&&5&$MAACD.LC$  &150&2&7&Yes &69&9\\
\hline\hline
Algo&$a_1$&$a_2$&$a_{TH}$&$T$& FP & TP  & \# seiz. & $\Delta_1$ & $\Delta_2$ \\ \hline
$TH$&1&0&1.75 &150&0&2&Not &14&-\\
$TH$&1&0&2&150&0&0&Not &  - &-\\
$TH$&1&0&2.25 &150&0&0&Not &  - &-\\
\hline
\end{tabular}
\end{center}
\caption{Results on patient $Chb23$.}
\label{Chb23}
\end{table}
}

{\small
\begin{table}
\begin{center}
\begin{tabular}{l r r r r r r r r r r r}
Algo&&$np$ & $Feat.$  & $T$ & FP & TP  & \# seiz. & $\Delta_1$ & $\Delta_2$ \\ \hline
$R-SVM$&&5&$Str.R1$ &150&0&18&Yes &52&21\\
$R-SVM$&&10&$MAACD.R1$  &150&0&17&Yes &57&31\\
$R-SVM$&&10&$MAACD.R1$  &200&0&16&Yes &52&26\\
\hline\hline
Algo&$a_1$&$a_2$&$a_{TH}$&$T$& FP & TP  & \# seiz. & $\Delta_1$ & $\Delta_2$ \\ \hline
$TH$&1&0&1.5 &150&0&17&Yes &47&21\\
$TH$&1&0&1.75 &200&0&17&Yes &47&21\\
$TH$&1&0&1.75 &300&0&17&Yes &47&21\\
\hline
\end{tabular}
\end{center}
\caption{Results on patient $Chb24$.}
\label{Chb24}
\end{table}
}

\section{Conclusions}\label{sec:conc}

Anticipating epileptic seizures is a very important open problem. 
Mainstream approach has investigated for many years on the analysis of EEG signal. 
It is currently understood that this phenomenon is not merely due to one single part of the brain, but rather to the (synchronized) interaction of an ensemble of brain parts.
Therefore, in order to analyze the synchronization patterns in the EEG signal, 
we have proposed a graph model of the brain interactions, in which we considered several synchronization measures.
Moreover, we propose an easily computable indicator function, called MAACD, to better capture the variations in the synchronization measures.

The data obtained in this manner are then used to identify the preictal state by means of two binary classification approaches.
To this aim, we have proposed a simple feature selection algorithm tailored for this specific application, and we also test the known Relief algorithm for feature selection.
Finally, we develop a simple linear classifier, again tailored on our application, and we also use a Support Vector Machine algorithm. 
Computational tests on real data from the ``CHB-MIT Scalp EEG Database" show that the simple and computationally viable approach proposed in this work 
is able to effectively detect the changes in the synchronization corresponding to the preictal state. 
Furthermore, although these two classifications techniques obtain comparable performances, 
the linear one yields a smaller number of false positives and a larger prediction time.

Future research directions include: $(i)$ the evaluation of different graph measures; $(ii)$ the development of more sophisticated methods for the computation of the thresholds in the linear classifier;  $(iii)$ testing the algorithms on  larger EEG scalp datasets.

\section*{Acknowledgements}
The research has been partially supported by the grant ``PANACEE'' (Prevision and analysis of brain activity in transitions: epilepsy and sleep)  of the Regione Toscana - PAR FAS  2007-2013 1.1.a.1.1.2 - B22I14000770002.

\end{document}